\documentclass[11pt,letterpaper]{article}
\usepackage{eacl2017}
\usepackage{times}
\usepackage{svg}
\usepackage{latexsym}
\usepackage{color}
\usepackage{graphicx}
\usepackage{times}
\usepackage{url}
\usepackage{color}
\usepackage{multirow}
\usepackage{arydshln}
\usepackage{mdframed}
\usepackage{amsmath}
\usepackage{amssymb}
\usepackage{mathtools}
\usepackage{booktabs}
\usepackage{enumitem}
\usepackage{multirow}
\usepackage{alltt}
\usepackage{mathabx}
\usepackage{mathtools}
\usepackage{subfig}
\usepackage{enumitem}
\newcommand{\ignore}[1]{}

\usepackage{cleveref}
\crefname{section}{§}{§§}
\Crefname{section}{§}{§§}

\eaclfinalcopy 

\title{Chains of Reasoning over Entities, Relations, and Text using \\ Recurrent Neural Networks}

\author{Rajarshi Das, Arvind Neelakantan, David Belanger, Andrew McCallum \\
  College of Information and Computer Sciences\\
  University of Massachusetts Amherst\\
  {\tt \{rajarshi, arvind, belanger, mccallum\}}@cs.umass.edu 
  }

\date{}

\begin{document}
\maketitle

\begin{abstract}
Our goal is to combine the rich multi-step inference of symbolic logical reasoning with the generalization capabilities of neural networks.  We are particularly interested in complex reasoning about entities and relations in text and large-scale knowledge bases (KBs). \newcite{neelakantan15} use RNNs to compose the distributed semantics of multi-hop paths in KBs; however for multiple reasons, the approach lacks accuracy and practicality. This paper proposes three significant modeling advances: (1) we learn to jointly reason about relations, \emph{entities, and entity-types}; (2) we use neural attention modeling to incorporate \emph{multiple paths}; (3) we learn to \emph{share strength in a single RNN} that represents logical composition across all relations. On a large-scale Freebase+ClueWeb prediction task, we achieve 25\% error reduction, and a 53\% error reduction on sparse relations due to shared strength. On chains of reasoning in WordNet we reduce error in mean quantile by 84\% versus previous state-of-the-art.\footnote{The code and data are available at https://rajarshd.github.io/ChainsofReasoning/}.
\end{abstract}

\section{Introduction} 
\label{sec:introduction}
\begin{table} 
\begin{tabular}{l l}\hline
i. place.birth$(a,b)$ $\leftarrow$ `$was\_born\_in$'$(a,x) \land$\\
`$commonly\_known\_as$'$(x,b)$\\
ii. location.contains$(a,b)$ $\leftarrow$ (nationality)$^{-1}(a,x) \land$\\
 place.birth$(x,b)$\\
iii. book.characters$(a,b)$ $\leftarrow $`$aka$'$(a,x) \land$\\
(theater.character.plays)$^{-1}(x,b)$\\
iv. cause.death$(a,b) \leftarrow $`$contracted$'$(a,b)$\\\hline

\end{tabular}
\begin{center}
\caption{Several highly probable clauses learnt by our model. The textual relations are shown in quotes and italicized. Our model has the ability to combine textual and schema relations. $r^{-1}$ is the inverse relation $r$, i.e. $r(a,b) \Leftrightarrow r^{-1}(b,a)$.}
\label{tab:eg1}
\end{center}
\end{table}

There is a rising interest in extending neural networks to perform more complex reasoning, formerly addressed only by symbolic and logical reasoning systems.  So far this work has mostly focused on small or synthetic data \cite{GrefenstetteTFDS,snli:emnlp2015,rocktaschel2016learning}. Our interest is primarily in reasoning about large knowledge bases (KBs) with diverse semantics, populated from text. One method for populating a KB from text (and for representing diverse semantics in the KB) is \emph{Universal Schema} \cite{USchema:13,pat:15}, which learns vector embeddings capturing the semantic positioning of relation types - the union of all input relation types, both from the schemas of multiple structured KBs, as well as expressions of relations in natural language text.

An important reason to populate a KB is to support not only look-up-style question answering, but reasoning on its entities and relations in order to make inferences not directly stored in the KB. KBs are often highly incomplete \cite{Min13distantsupervision}, and reasoning can fill in these missing facts. The ``matrix completion'' mechanism that underlies the common implementation of Universal Schema can thus be seen as a simple type of reasoning, as can other work in tensor factorization \cite{Nickel,Bordes,Socher}. However these methods can be understood as operating on single pieces of evidence: for example, inferring that Microsoft--{\it located-in}--Seattle implies Microsoft--{\sl HQ-in}--Seattle.
\begin{figure*}[ht]
  \subfloat[]{\includegraphics[width=0.60\textwidth]{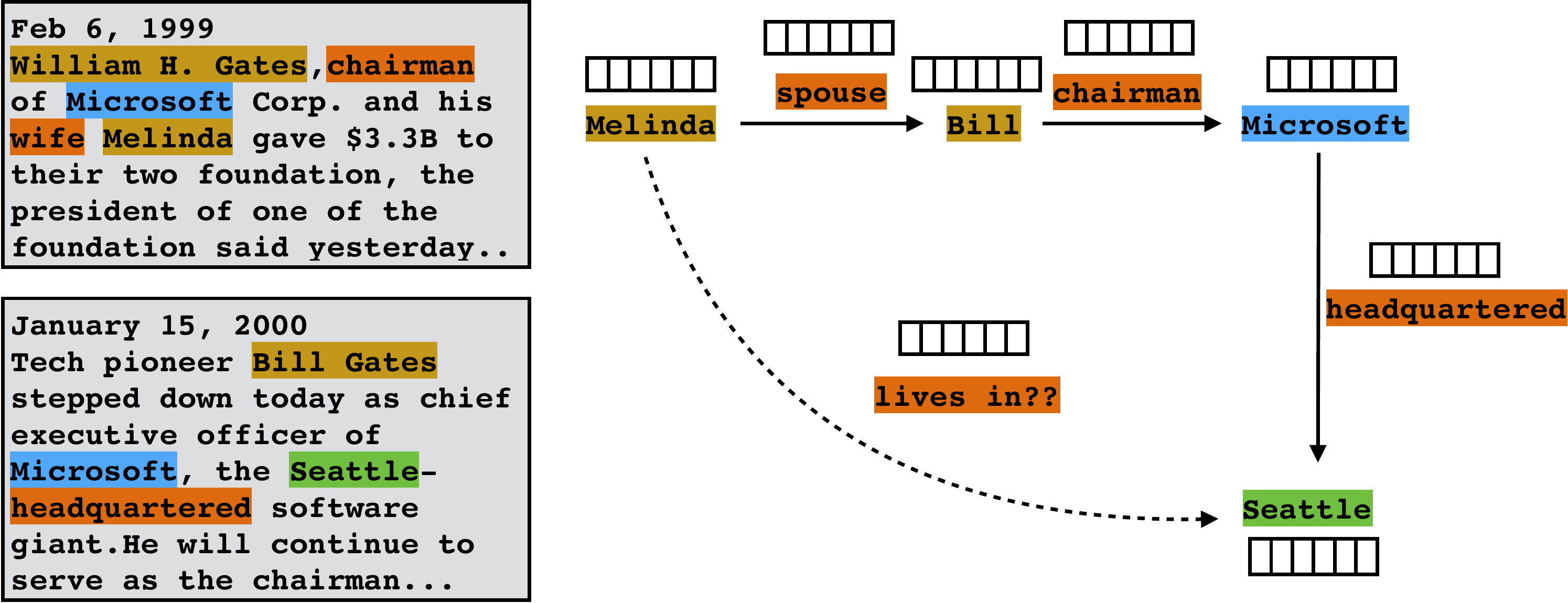}\label{fig:multi_hop_inference}}
  \subfloat[]{\includegraphics[width=0.45\textwidth]{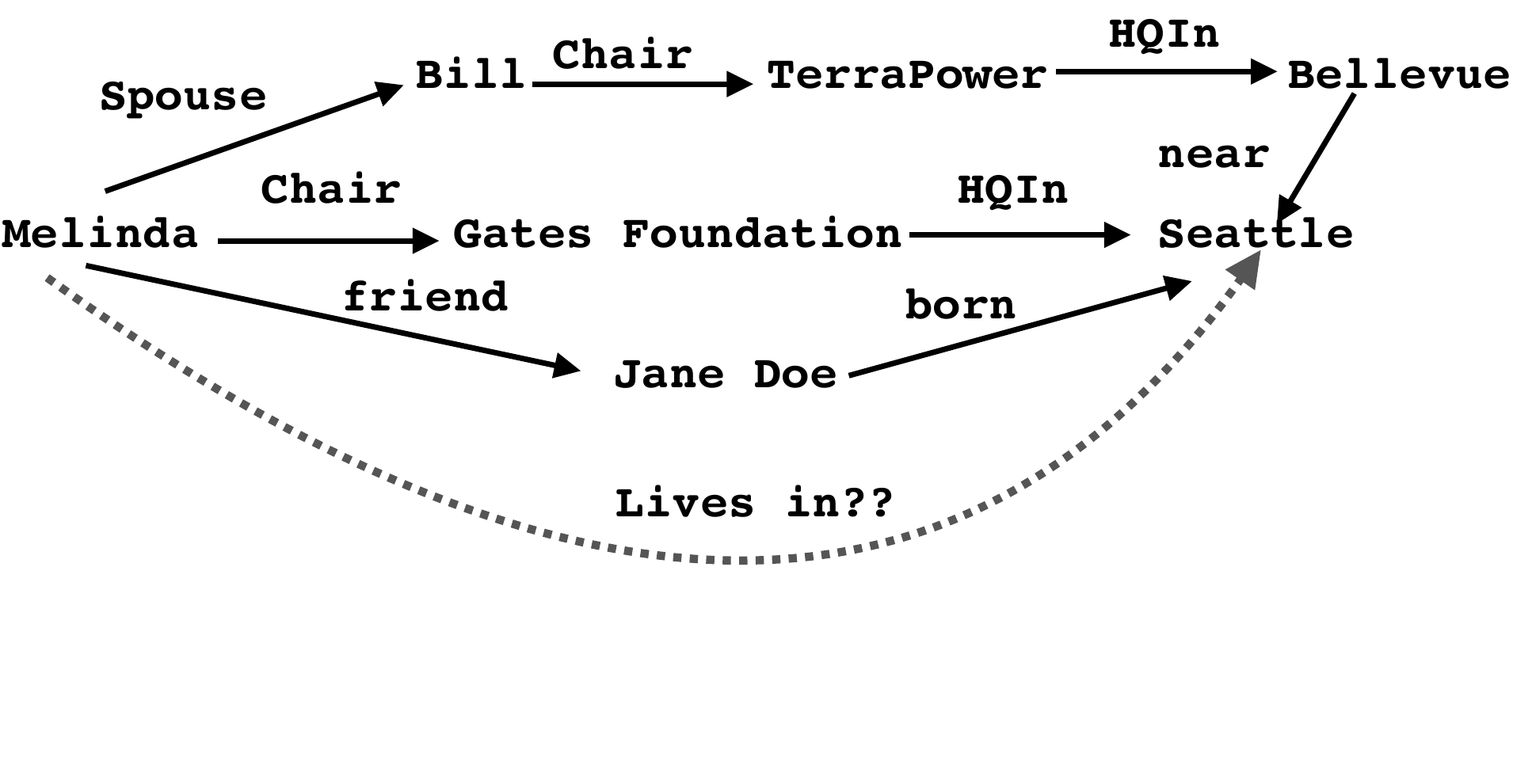}\label{fig:paths_melinda_seattle}}
  \caption{The nodes in the knowledge graphs represent entities and the labeled edges represent relations. (a) A path between `Melinda' and `Seattle' combining relations from two different documents. (b) There are multiple paths between entities in a knowledge graph. The top two paths are predictive of the fact that Melinda may `\textit{live in}' Seattle, but the bottom (fictitious) path isn't.}
  \end{figure*}

A highly desirable, richer style of reasoning makes inferences from Horn clauses that form multi-hop paths containing three or more entities in the KB's entity-relation graph.  For example, we may have no evidence directly linking Melinda Gates and Seattle, however, we may infer with some likelihood that Melinda--{\it lives-in}--Seattle, by observing that the KB contains the path Melinda--{\it spouse}--Bill--{\it chairman}--Microsoft--{\sl HQ-in}--Seattle (Fig.~\ref{fig:multi_hop_inference}).

 Symbolic rules of this form are learned by the Path Ranking Algorithm (PRA) \cite{Lao:2011}.  Dramatic improvement in generalization can be obtained by reasoning about paths, not in terms of relation-symbols, but Universal Schema style relation-vector-embeddings.  This is done by \newcite{neelakantan15}, where RNNs semantically compose the per-edge relation embeddings along an arbitrary-length path, and output a vector embedding representing the inferred relation between the two entities at the end-points of the path. This approach thus represents a key example of complex reasoning over Horn clause chains using neural networks. However, for multiple reasons detailed below it is inaccurate and impractical.

 This paper presents multiple modeling advances that significantly increase the accuracy and practicality of RNN-based reasoning on Horn clause chains in large-scale KBs.  (1) Previous work, including \cite{Lao:2011,neelakantan15,gu2015} reason about chains of relations, but not the entities that form the nodes of the path. Ignoring entities and entity-types leads to frequent errors, such as inferring that Yankee Stadium serves as a transportation hub for NY state. In our work, we jointly learn and reason about relation-types, entities, and entity-types. (2) The same previous work takes only a single path as evidence in inferring new predictions. However, as shown in Figure~\ref{fig:paths_melinda_seattle}, multiple paths can provide evidence for a prediction. In our work, we use neural attention mechanisms to reason about multiple paths. We use a novel pooling function which does soft attention during \emph{gradient step} and find it to work better.
 (3) The most problematic impracticality of the above previous work\footnote{with exception of \cite{gu2015}} for application to KBs with broad semantics is their requirement to train a separate model for each relation-type to be predicted. In contrast, we train a single, high-capacity RNN that can predict all relation types. In addition to efficiency advantages, our approach significantly increases accuracy because the multi-task nature of the training shares strength in the common RNN parameters.

 We evaluate our new approach on a large scale dataset of Freebase entities, relations and ClueWeb text\ignore{(over 3m entity pairs, 23k textual relation patterns)}. In comparison with the previous best on this data, we achieve an error reduction of 25\% in mean average precision (MAP). In an experiment specially designed to explore the benefits of sharing strength with a single RNN, we show a 54\% error reduction in relations that are available only sparsely at training time. We also evaluate on a second data set, chains of reasoning in WordNet. In comparison with previous state-of-the-art \cite{gu2015} our model achieves a 84\% reduction in error in mean quantile.

\section{Background} 
\label{sec:background}
\label{sub:rnn_model}
\begin{figure*}[ht]
  \centering
    \includegraphics[width=0.8\textwidth]{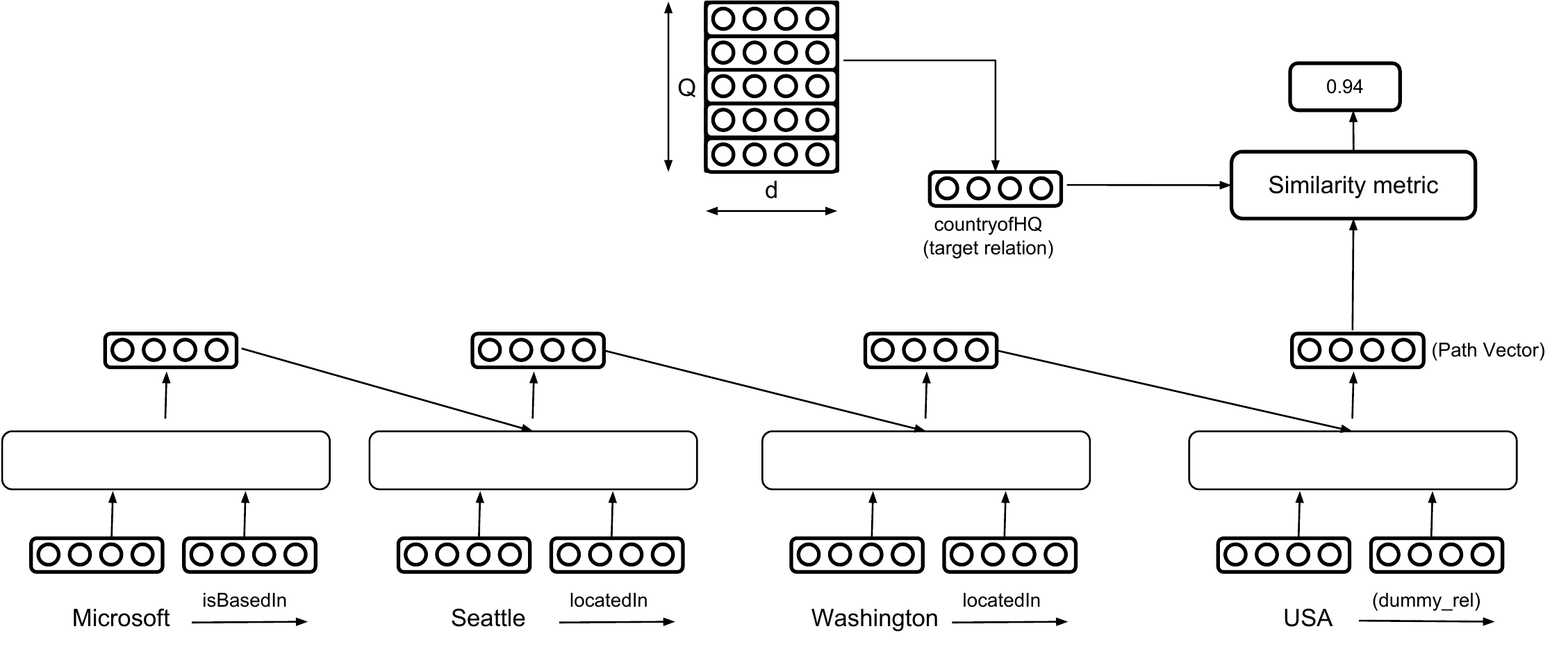}
    \caption{At each step, the RNN consumes both entity and relation vectors of the path. The entity representation can be obtained from its types. The path vector $\mathbf{y_{\pi}}$ is the last hidden state. The parameters of the RNN and relation embeddings are shared across all query relations. The dot product between the final representation of the path and the query relation gives a confidence score, with higher scores indicating that the query relation exists between the entity pair.}
\label{fig:RNNModel}
\end{figure*}

In this section, we introduce the compositional model (Path-RNN) of \newcite{neelakantan15}. The Path-RNN model takes in input a path between two entities and infers new relations between them. Reasoning is performed non-atomically about conjunctions of relations in an arbitrary length path by composing them with a recurrent neural network (RNN). The representation of the path is given by the last hidden state of the RNN obtained after processing all the relations in the path.

Let $\left(e_s, e_t\right)$ be an entity pair and $\mathcal{S}$ denote the set of paths between them. The set $\mathcal{S}$ is obtained by doing random walks in the knowledge graph starting from $e_s$ till $e_t$. Let $\pi=\{e_s, r_1, e_1, r_2, \ldots, r_k, e_t\} \in \mathcal{S}$ denote a path between $\left(e_s, e_t\right)$. The length of a path is the number of relations in it, hence, $(\mathrm{len}(\pi) = k)$. Let $\mathbf{y_{r_t}} \in \mathbb{R}^{d} $ denote the vector representation of $r_t$. The Path-RNN model combines all the \textit{relations} in $\pi$ sequentially using a RNN with an intermediate representation $\mathbf{h_{t} \in \mathbb{R}^{h}}$ at step $t$ given by
\begin{equation}
  \mathbf{h_t} = f(\mathbf{W^{r}_{hh}}\mathbf{h_{t-1}} + \mathbf{W^{r}_{ih}}\mathbf{y^{r}_{r_t}}).
  \label{eq:ht}
\end{equation}
$\mathbf{W^{r}_{hh}} \in \mathbb{R}^{h \times h}$ and $\mathbf{W^{r}_{ih}} \in \mathbb{R}^{d \times h}$ are the parameters of the RNN. Here $r$ denotes the query relation. Path-RNN has a specialized model for predicting each query relation $r$, with separate parameters $(\mathbf{y^{r}_{r_t}}, \mathbf{W^{r}_{hh}}, \mathbf{W^{r}_{ih}})$ for each $r$. $f$ is the sigmoid function. The vector representation of path $\pi$ $(\mathbf{y_{\pi}})$ is the last hidden state $\mathbf{h_k}$. The similarity of $\mathbf{y_{\pi}}$ with the query relation vector $\mathbf{y_{r}}$ is computed as the dot product between them:
\begin{equation}
	\mathrm{score(\pi, r)} = \mathbf{y_{\pi}} \cdot \mathbf{y_{r}}
  \label{eq:score}	
\end{equation} 
Pairs of entities may have several paths connecting them in the knowledge graph (Figure~\ref{fig:paths_melinda_seattle}). Path-RNN computes the probability that the entity pair $(e_{s}, e_t)$ participates in the query relation $(r)$ by, 
\begin{align}
  \mathbb{P}(r|e_s, e_t) & = &\max \sigma(\mathrm{score}(\pi, r)) ,\forall \pi \in \mathcal{S}
  \label{eq:p_rnn}
\end{align}
where $\sigma$ is the \textit{sigmoid} function.

Path-RNN and other models such as the Path Ranking Algorithm (PRA) and its extensions \cite{Lao:2011,Lao:2012,pra_recent,vector_pra} makes it impractical to be used in downstream applications, since it requires training and maintaining a model for each relation type. Moreover, parameters are not shared across multiple target relation types leading to large number of parameters to be learned from the training data.

In \eqref{eq:p_rnn}, the Path-RNN model selects the maximum scoring path between an entity pair to make a prediction, possibly ignoring evidence from other important paths. Not only is this a waste of computation (since we have to compute a forward pass for all the paths anyway), but also the relations in all other paths do not get any gradients updates during training as the max operation returns zero gradient for all other paths except the maximum scoring one. This is especially ineffective during the initial stages of the training since the maximum probable path will be random.

The Path-RNN model and other multi-hop relation extraction approaches (such as \newcite{gu2015}) ignore the entities in the path. Consider the following paths JFK--{\it locatedIn}--NYC--{\it locatedIn}--NY and Yankee Stadium--{\it locatedIn}--NYC--{\it locatedIn}--NY. To predict the $airport\_serves$ relation, the Path-RNN model assigns the same scores to both the paths even though the first path should be ranked higher. This is because the model does not have information about the entities and just uses the relations in the path for prediction.


\section{Modeling Approach} 
\label{sec:single_model}
\subsection{Shared Parameter Architecture} 
\label{sub:one_model}
 Previous section discussed the problems associated with per-relation modeling approaches. In response, we \emph{share} the relation type representation and the composition matrices of the RNN across all target relations enabling lesser number of parameters for the same training data. We refer to this model as \emph{Single-Model}. Note that this is just \emph{multi-task learning} \cite{Caruana1997} among prediction of target relation types with an underlying shared parameter architecture.
The RNN hidden state in \eqref{eq:ht} is now given by:
\begin{equation}
	\mathbf{h_t} = f(\mathbf{W_{hh}h_{t-1}} + \mathbf{W_{ih}y_{r_t}}).
	\label{eq:single_model_hidden_state_eq}
\end{equation}
Readers should take note that the parameters here are independent of each target relation $r$.\ignore{ Also $f$ is a ReLU function \cite{relu} instead of sigmoid.}
\subsection*{Model Training} 
\label{ssub:model_training}
We train the model using existing observed facts (triples) in the KB as positive examples and unobserved facts as negative examples. Let $\mathcal{R} = \{\gamma_1, \gamma_2,\ldots,\gamma_{n}\}$ denote the set of all query relation types that we train for. Let $\Delta^{+}_{\mathcal{R}}, \Delta^{-}_{\mathcal{R}}$ denote the set of positive and negative triples for \textit{all} the relation types in $\mathcal{R}$. The parameters of the model are trained to minimize the negative $\log$-likelihood of the data.
\begin{align}
	L(\Theta, \Delta^{+}_{\mathcal{R}}, \Delta^{-}_{\mathcal{R}} ) = -\frac{1}{M}\sum_{e_s,e_t,r \in \Delta^{+}_{\mathcal{R}}}\log \mathbb{P}(r|e_s, e_t) \nonumber\\ + \sum_{\hat{e}_s,\hat{e}_t,\hat{r} \in \Delta^{-}_{\mathcal{R}}}\log(\ 1 - \mathbb{P}(\hat{r}|\hat{e}_s, \hat{e}_t))
\label{loss_function}
\end{align}
Here $M$ is the total number of training examples and $\Theta$ denotes the set of all parameters of the model (lookup table of embeddings (shared) and parameters of the RNN (shared)). It should be noted that the Path-RNN model has a separate loss function for each relation $r \in \mathcal{R}$ which depends only on the relevant subset of the data.

\subsection{Score Pooling} 
\label{sub:score_pooling}
In this section, we introduce new methods of score pooling that takes into account multiple paths between an entity pair. Let $\left\{s_1, s_2,\ldots,s_N\right\}$ be the similarity scores (Equation \ref{eq:score}) for $N$ paths connecting an entity pair $(e_s, e_t)$. The probability for entity pair $(e_s, e_t)$ to participate in relation $r$ (Equation \ref{eq:p_rnn}) is now given by,
\begin{enumerate}
	\item Top-($k$): A straightforward extension of the `max' approach in which we average the top $k$ scoring paths. Let $\mathcal{K}$ denote the indices of top-$k$ scoring paths. 
	\begin{align}
		\mathbb{P}(r|e_s, e_t) = \sigma(\frac{1}{k}\sum_{j} s_j), \forall j \in \mathcal{K} \nonumber
	\end{align} 
	\item Average: Here, the final score is the average of scores of all the paths.
	\begin{align}
		\mathbb{P}(r|e_s, e_t) = \sigma(\frac{1}{N}\sum_{i=1}^{N}s_i)\nonumber
		\label{eq:p_lse}
	\end{align} 	
	\item LogSumExp: In this approach the pooling layer is a smooth approximation to the `max' function - LogSumExp (LSE). Given a vector of scores, the LSE is calculated as
	\begin{align}
		\mathrm{LSE}(s_1, s_2,\ldots, s_n) = \log(\sum_{i}\exp(s_{i}))\nonumber
	\end{align}
	and hence the probability of the triple is,
	\begin{align}
		\mathbb{P}(r|e_1, e_2) = \sigma(\mathrm{LSE}(s_1, s_2,\ldots,s_n))\nonumber
	\end{align}
\end{enumerate} 
The average and the LSE pooling functions apply non-zero weights to \emph{all} the paths during inference. However only a few paths between an entity pair is predictive of a query relation. LSE has another desirable property since $\frac{\partial \mathrm{LSE}}{\partial s_i} = \frac{\exp(s_i)}{\sum_{i}\exp(s_i)}$. This means that during the back-propagation step, every path will receive a share of the gradient proportional to its score and hence this is a kind of novel neural attention during the gradient step. In contrast, for averaging, every path will receive equal $(\frac{1}{N})$ share of the gradient. Top-($k$) (similar to max) receives sparse gradients.
\subsection{Incorporating Entities} 
\label{sub:incorporating_selectional_preferences_model}
A straightforward way of incorporating entities is to include entity representations (along with relations) as input to the RNN. Learning separate representations of entity, however has some disadvantages. The distribution of entity occurrence is heavy tailed and hence it is hard to learn good representations of rarely occurring entities. To alleviate this problem, we use the entity types present in the KB as described below.

Most KBs have annotated types for entities and each entity can have multiple types. For example, Melinda Gates has types such as \textit{CEO, Duke University Alumni, Philanthropist, American Citizen} etc. We obtain the entity representation by a simple addition of the entity type representations. The entity type representations are learned during training. We limit the number of entity types for an entity to 7 most frequently occurring types in the KB. Let $\mathbf{y_{e_t}} \in \mathbb{R}^{m}$ denote the representation of entity $e_t$, then \ref{eq:single_model_hidden_state_eq} now becomes
\begin{equation}
	\mathbf{h_t} = f(\mathbf{W_{hh}h_{t-1}} + \mathbf{W_{ih}y_{r_t}} + \mathbf{W_{eh}y_{e_t}})
	\label{eq:single_model_hidden_state_entities_eq}
\end{equation}
$\mathbf{W_{eh}} \in \mathbb{R}^{m \times h}$ is the new parameter matrix for projecting the entity representation. Figure \ref{fig:RNNModel} shows our model with an example path between entities (\textit{Microsoft, USA}) with \textit{countryOfHQ} (country of head-quarters) as the query relation.

\section{Related Work} 
\label{sec:related_work}
Two early works on extracting clauses and reasoning over paths are SHERLOCK \cite{Schoenmackers:2010} and the Path Ranking Algorithm (PRA) \cite{Lao:2011}. SHERLOCK extracts purely symbolic clauses by exhaustively exploring relational paths of increasing length. PRA replaces exhaustive search by random walks. Observed paths are used as features for a per-target-relation binary classifier. \newcite{Lao:2012} extend PRA by augmenting KB-schema relations with observed text patterns. However, these methods do not generalize well to millions of distinct paths obtained from random exploration of the KB, since each unique path is treated as a singleton, where no commonalities between paths are modeled. In response, pre-trained vector representations have been used in PRA to tackle the feature explosion~\cite{pra_recent,vector_pra} but still rely on a classifier using atomic path features.\newcite{yang} also extract horn rules but they restrict it to a length of 3 and are restricted to schema types. \newcite{wenyuan} show improvements in relation extraction by incorporating sentences which contain one entity by connecting them through a path. 

\newcite{gu2015} introduce new compositional techniques by modeling additive and multiplicative interactions between relation matrices in the path. However they model only a \textit{single} path between an entity pair in-contrast to our ability to consider multiple paths. \newcite{toutanova16} improves upon them by additionally modeling the intermediate entities in the path and modeling multiple paths. However, in their approach they have to store scores for intermediate path length for \textit{all} entity pairs, making it prohibitive to be used in our setting where we have more than 3M entity pairs. They also model entities as just a scalar weight whereas we learn both entity and type representations. Lastly it has been shown by \newcite{neelakantan15} that non-linear composition function out-performs linear functions (as used by them) for relation extraction tasks.

The performance of relation extraction methods have been improved by incorporating entity types for their candidate entities, both in sentence level \cite{droth,jnt:akbc13} and KB relation extraction \cite{trescal}, and in learning entailment rules \cite{Berant:2011}. \newcite{serban2016generating} use RNNs to generate factoid question from Freebase.

\section{Results} 
\label{sec:experiments}

\subsection*{Data and Experimental Setup} 
\label{sub:data_and_experimental_setup}

\label{sub:data}
\begin{table}
\begin{center}
\small
\begin{tabular}{l c}\hline
Stats & \#\\\hline
\# Freebase relation types & 27,791\\
\# textual relation types & 23,599 \\
\# query relation types & 46\\
\# entity pairs & 3.22M\\
\# unique entity types & 2218\\
Avg. path length & 4.7\\
Max path length & 7\\
Total \# paths & 191M\\\hline

\end{tabular}
\end{center}
\begin{center}
\caption{Statistics of the dataset.}
\label{tab:data}
\end{center}
\end{table}
We apply our models to the dataset released by \newcite{neelakantan15}, which is a subset of Freebase enriched with information from ClueWeb. The dataset is comprised of a set of triples (\textit{$e_1$, $r$, $e_2$}) and also the set of paths connecting the entity pair (\textit{$e_1$,$e_2$}) in the knowledge graph. The triples extracted from ClueWeb consists of sentences that contained entities linked to Freebase \cite{clue_web}. The phrase between the two entities in the sentence forms the relation type. To limit the number of textual relations, we retain the two following words after the first entity and two words before the second entity.\ignore{However, the paths in the dataset has the entity information missing from theirs and only contains the relation types occurring in them. For example, consider the path $\text{Satya Nadella}\xrightarrow[]{\text{ceo of}} \text{Microsoft} \xrightarrow[]{\text{locatedIn}} \text{Seattle}$.The original dataset had the entities in-between such as `$\text{Microsoft}$' and `$\text{Seattle}$' missing from it. We augment the dataset with the entities present in the paths.\ignore{To gather the entities, we do a depth first traversal in the Freebase knowledge graph starting from the first entity of the entity pair and following the relation types until we reach the last entity of the pair. In cases of one-to-many relations  we choose the next entity to be traversed at random.}} We also collect the entity type information from Freebase. Table~\ref{tab:data} summarizes some important statistics. For the PathQA experiment, we use the same train/dev/test split of WordNet dataset released by \newcite{gu2015} and hence our results are directly comparable to them. The WordNet dataset has just 22 relation types and 38194 entities which is order of magnitudes less than the dataset we use for relation extraction tasks.
\begin{center}
\begin{table*}[ht]
\begin{center}
\small
\begin{tabular}{c c c c}\hline
Model & Performance (\%MAP) & Pooling \\\hline
Single-Model & 68.77 & Max\\
Single-Model & 55.80 & Avg.\\
Single-Model & 68.20 & Top($k$)\\
Single-Model & \textbf{70.11} & LogSumExp\\\hline
PRA & 64.43 & n/a \\
PRA + Bigram & 64.93 & n/a\\
Path-RNN & 65.23 & Max\\
Path-RNN & 68.43 & LogSumExp\\
Single-Model & \textbf{70.11} & LogSumExp\\\hline
PRA + Types & 64.18 & n/a\\
Single-Model & 70.11 & LogSumExp\\
Single-Model + Entity & 71.74 & LogSumExp\\
Single-Model + Types & \textbf{73.26} & LogSumExp\\
Single-Model + Entity + Types & 72.22 & LogSumExp\\\hline
\end{tabular}
\caption{The first section shows the effectiveness of LogSumExp as the score aggregation function. The next section compares performance with existing multi-hop approaches and the last section shows the performance achieved using joint reasoning with entities and types.}
\label{tab:all_results}
\end{center}
\end{table*}
\end{center}
The dimension of the relation type representations and the RNN hidden states are $d, h=250$ and the entity and type embeddings have $m=50$ dimensions. The Path-RNN model has sigmoid units as their activation function. However, we found rectifier units (ReLU) to work much better~\cite{LeJH15}\footnote{even when compared to LSTMs (73.2 vs 72.4 in MAP)}	.\ignore{We had a batch size of ($B = 32$) for our relation extraction experiment and ($B_{t} = 40$) for entity-type prediction experiments.} For the path-query experiment, the dimension of entity, relation embeddings and hidden units are set to 100 (as used by \newcite{gu2015}). As our evaluation metric, we use the average precision (AP) to score the ranked list of entity pairs. The MAP score is the mean AP across all query relations. AP is a strict metric since it penalizes when an incorrect entity is ranked higher above correct entities. Also MAP approximates the area under the Precision Recall curve \cite{Manning:2008}. We use Adam~\cite{adam} for optimization for all our experiments with the default hyperparameter settings (learning rate = 1$e^{-3}$, $\beta_{1}=0.9$, $\beta_{2}=0.999$, $\epsilon=1e^{-8}$). Statistical significance for scores reported in Table \ref{tab:all_results} were done with a paired-$t$ test.



\subsection{Effect of Pooling Techniques} 
\label{sub:effect_of_pooling_techniques}
Section 1 of Table \ref{tab:all_results} shows the effect of the various pooling techniques presented in section \ref{sub:score_pooling}. It is encouraging to see that \textit{LogSumExp} gives the best results. This demonstrates the importance of considering information from all the paths. However, Avg.\ pooling performs the worst, which shows that it is also important to weigh the paths scores according to their values.
\begin{figure}[t]
  \centering
    \includegraphics[width=0.40\textwidth]{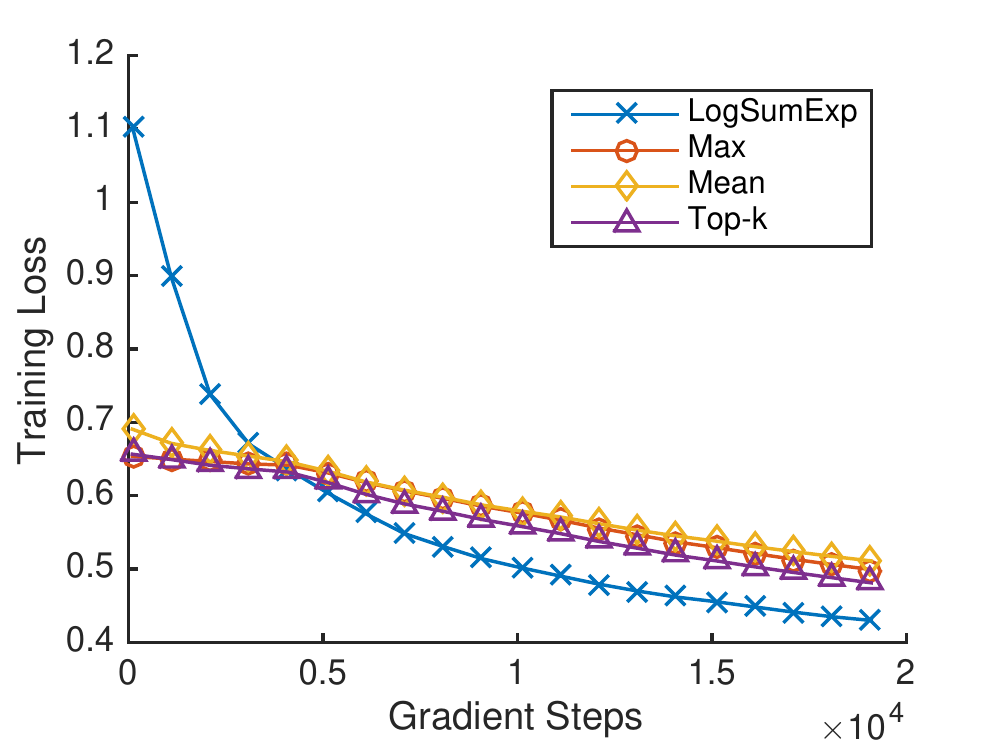}
    \caption{Comparison of the training loss w.r.t gradient update steps of various pooling methods. The loss of LogSumExp decreases the fastest among all pooling methods and hence leads to faster training.}
    \label{fig:train_loss_plot}
 \end{figure}
 Figure~\ref{fig:train_loss_plot} plots the training loss w.r.t gradient update step. Due to non-zero gradient updates for all the paths, the LogSumExp pooling strategy leads to faster training vs. max pooling, which has sparse gradients. This is especially relevant during the early stages of training, where the argmax path is essentially a random guess. The scores of max and LSE pooling are significant with ($p < 0.02$).

\subsection{Comparison with multi-hop models} 
\label{sub:comparison_with_multi_hop_models}
We next compare the performance of the Single-Model with two other multi-hop models - Path-RNN and PRA\cite{Lao:2011}. Both of these approaches train an individual model for each query relation.\ignore{ model, so this experimental setup tests whether having a single-model with shared parameters is a better modeling technique.} We also experiment with another extension of PRA that adds bigram features (PRA + Bigram). Additionally, we run an experiment replacing the max-pooling of Path-RNN with LogSumExp. The results are shown in the second section of Table~\ref{tab:all_results}. It is not surprising to see that the Single-Model, which leverages parameter sharing improves performance. It is also encouraging to see that LogSumExp makes the Path-RNN baseline stronger. The scores of Path-RNN (with LSE) and Single-Model are significant with ($p<0.005$).

\subsection{Effect of Incorporating Entities} 
\label{sub:incorporating_selectional_preferences}
Next, we provide quantitative results supporting our claim that modeling the entities along a KB path can improve reasoning performance. The last section of Table~\ref{tab:all_results} lists the performance gain obtained by injecting information about entities. We achieve the best performance when we represent entities as a function of their annotated types in Freebase (Single-Model + Types) $(p<0.005)$. In comparison, learning separate representations of entities (Single-Model + Entities) gives slightly worse performance. This is primarily because we encounter many new entities during test time, for which our model does not have a learned representation. However the relatively limited number of entity types helps us overcome the problem of representing unseen entities. \ignore{We also experimented with a scenario where an entity is represented as the sum of learned representation of itself and from its types (Single-Model + Entity + Types). This straightforward extension of equations~\ref{eq:E_eq} and~\ref{eq:T_eq} gave slightly better performance than learning a single representation of entities, but it was lower than representing entities just with its types.} We also extend PRA to include entity type information (PRA + Types), but this did not yield significant improvements.

\subsection{Performance in Limited Data Regime} 
\label{sub:performance_in_limited_data_regime_}
In constructing our dataset, we selected query relations with reasonable amounts of data. However, for many important applications we have very limited data. To simulate this common scenario, we create a new dataset by randomly selecting 23 out of 46 relations and removing all but 1\% of the positive and negative triples previously used for training. Effectively, the difference between Path-RNN and Single-Model is that Single-Model does multitask learning, since it shares parameters for different target relation types. Therefore, we expect it to outperform Path-RNN on this small dataset, since this multitask learning provides additional regularization. We also experiment with an extension of Single-Model where we introduce an additional task for multitask learning, where we seek to predict annotated types for entities. Here, parameters for the entity type embeddings are shared with the Single-Model. Supervision for this task is provided by the entity type annotation in the KB.  We train with a Bayesian Personalized Ranking loss of \newcite{bpr}. The results are listed in Table \ref{tab:mtl_results}. With Single-Model there is a clear jump in performance as we expect. The additional multitask training with types gives a very incremental gain.

\begin{table}
\begin{center}
\begin{tabular}{c c}\hline
Model & Performance (\%MAP)\\\hline
Path-RNN & 22.06\\
Single-Model & 63.33\\
Single-Model + MTL & 64.81\\\hline
\end{tabular}
\caption{Model performance when trained with a small fraction of the data.}
\label{tab:mtl_results}
\end{center}
\end{table}

\begin{center}
\begin{table*}[ht] 
\begin{tabular}{c c c c }\hline
Horn Clause (Body) & Without Entities & With Entities & Universal \\\hline
location.contains$(x,a)$ $\land$ location.contains$(a,y)$ & 0.9149 & 0.949 & Y\\
(person.nationality)$^{-1}(x,a)$ $\land$ place.birth$(a,y)$ & 0.7702 & 0.9256 & N\\\hline
\end{tabular}
\begin{center}
\small
\caption{Body of two clauses both of which are predictive of $\text{location.contains}(x,y)$. First fact is universally true but the truth value of the second clause depends on the value of the entities in the clause. The model without entity parameters cannot discriminate this and outputs a lower overall confidence score.}
\label{tab:existential_quants}
\end{center}
\end{table*}
\end{center}

\subsection{Answering Path Queries} 
\label{sub:answering_path_queries_exp}
\newcite{gu2015} introduce a task of answering questions formulated as path traversals in a KB. Unlike binary fact prediction, to answer a path query, the model needs to find the set of correct target entities `$t$' that can be reached by starting from an initial entity `$s$' and then traversing the path `$p$'. They model additive and multiplicative interactions of relations in the path.\ignore{For example, in their compositional Bilinear-Diag model, the score of a triple $(s,p,t)$ is $x_{s}^{\top}W_{r_{1}}\cdot W_{r_{2}}\ldots W_{r_{n}}x_{t}$ and for compositional Trans-E the score is $-||x_{s}+w_{r_1}+\ldots+w_{r_n}-x_{t}||_{2}^{2}$ where $W_{r_{i}}$ and $w_{r_i}$ represents a diagonal matrix and vector associated with relation $r_i$ respectively.} It should be noted that the compositional Trans-E and Bilinear-diag have comparable number of parameters to our model since they also represent relations as vectors, however the Bilinear model learns a dense square \emph{matrix} for each relation and hence has a lot more number of parameters. Hence, we compare with Trans-E and Bilinear-diag models. Bilinear-diag has also been shown to outperform Bilinear models \cite{yang}.

Instead of combining relations using simple additions and multiplications, we propose to combine the intermediate hidden representations $h_i$ obtained from a RNN (via \eqref{eq:single_model_hidden_state_eq}) after consuming relation $r_i$ at each step. Let $\mathbf{h}$ denote the sum of all intermediate representations $h_{i}$. The score of a triple $(s,p,t)$ by our model is given by $x_{s}^{\top}\mathrm{diag}(\mathbf{h})x_{t}$ where $\mathrm{diag}(\mathbf{h})$ represents a diagonal matrix with vector $\mathbf{h}$ as its diagonal elements.

We compare to the results reported by \newcite{gu2015} on the WordNet dataset. It should be noted that the dataset is fairly small with just 22 relation types and an average path length of 3.07. More importantly, there are only few \textit{unseen paths} during test time and only \textit{one} path between an entity pair, suggesting that this dataset is not an ideal test bed for compositional neural models. The results are shown in table \ref{tab:path_qa_results}. Mean Quantile(MQ) is the fraction of incorrect entities which have been scored lower than the correct entity. Our model achieves a 84\% reduction in error when compared to their best model.
\begin{table}
\begin{center}
\begin{tabular}{c c}\hline
Model & MQ\\\hline
Comp. Bilinear Diag & 90.4\\
Comp. Trans-E & 93.3 \\
Our Model & \textbf{98.94}\\\hline
\end{tabular}
\caption{Performance on path queries in WordNet.}
\label{tab:path_qa_results}
\end{center}
\end{table}

\section{Qualitative Analysis} 
\label{sec:qualitative_analysis}
\textbf{Entities as Existential Quantifiers}: 
Table~\ref{tab:existential_quants} shows the body of two horn clauses. Both the clauses are predictive of the fact $\text{location.contains}(x,b)$. The first clause is \emph{universally} true irrespective of the entities present in the chain (transitive property). However the value of the second clause is only true \emph{conditional} on the instantiations of the entities. The score of the Path-RNN model is independent of the entity values, whereas our model outputs a different score based on the entities in the chain. We average the scores across entities, which are connected through this path and for which the relation holds in column 3 (With Entities).

For the first clause, which is independent of entities, both models predict a high score. However for the second clause, the model without entity information predicts a lower score because this path is seen in both positive and negative training examples and the model cannot condition on the entities to learn to discriminate. However our model predicts the true relations with high confidence. This is a step towards the capturing existential quantification for logical inference in vector space.

\begin{figure}[t]
  \centering
    \includegraphics[width=0.30\textwidth]{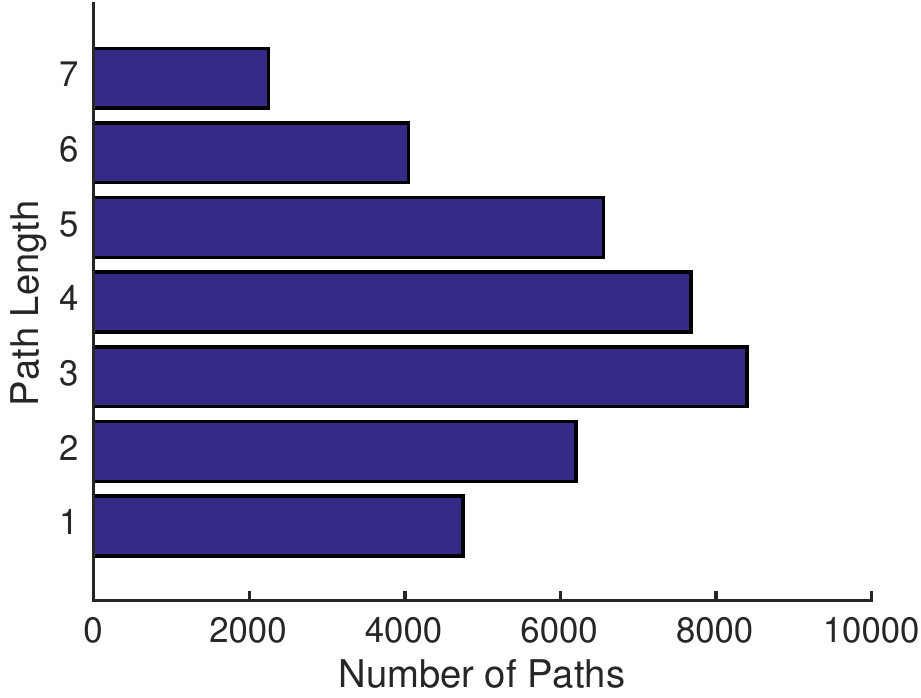}
    \caption{Length distribution of top-scoring paths}
    \label{fig:length_dist}
 \end{figure}

\noindent\textbf{Length of Clauses}: Figure \ref{fig:length_dist} shows the length distribution of top scoring paths in the test set. The distribution peaks at lengths$=\{3,4,5\}$, suggesting that previous approaches \cite{yang} which restrict the length to 3 might limit performance and generalizability.

\noindent\textbf{Limitation}: A major limitation of our model is inability to handle long textual patterns because of sparsity. Compositional approaches for modeling text \cite{toutanova:15,pat:15} are a right step in this direction and we leave this as future work.






\section{Conclusion} 
\label{sec:conclusion}
This paper introduces a single high capacity RNN model which allows chains of reasoning across multiple relation types. It leverages information from the intermediate entities present in the path between an entity pair and mitigates the problem of unseen entities by representing them as a function of their annotated types. We also demonstrate that pooling evidence across multiple paths improves both training speed and accuracy. Finally, we also address the problem of reasoning about infrequently occurring relations and show significant performance gains via multitasking.\ignore{with an auxiliary task of entity type prediction.}
\newpage
\bibliography{eacl2017}
\bibliographystyle{eacl2017}
\end{document}